\documentclass[11pt,a4paper]{article}
\usepackage[hyperref]{acl2020}
\usepackage{times}
\usepackage{latexsym}

\usepackage{graphicx}

\usepackage{microtype}

\aclfinalcopy 

\title{An Efficient Approach for Machine Translation on \\Low-resource Languages: A Case Study in Vietnamese-Chinese}

\author{Tran Ngoc Son $^*$, Nguyen Anh Tu $^*$, Nguyen Minh Tri  \\
Samsung SDS R\&D Center, Hanoi, Vietnam\\
 \texttt{\{tn.son,na.tu,nm.tri\}@gsamsung.com;}
} 

\begin{document}
\maketitle
\def\thefootnote{*}\footnotetext{These authors contributed equally to this work}\def\thefootnote{\arabic{footnote}}
\begin{abstract}
Despite the rise of recent neural networks in machine translation, those networks do not work well if the training data is insufficient. In this paper, we proposed an approach for machine translation in low-resource languages such as Vietnamese-Chinese. Our proposed method leveraged the power of the multilingual pre-trained language model (mBART) and both Vietnamese and Chinese monolingual corpus. Firstly, we built an early bird machine translation model using the bilingual training dataset. Secondly, we used TF-IDF technique to select sentences from the monolingual corpus which are the most related to domains of the parallel dataset. Finally, the first model was used to synthesize the augmented training data from the selected monolingual corpus for the translation model. Our proposed scheme showed that it outperformed 8\% compared to the transformer model. the augmented dataset also pushed the model performance.
\end{abstract}
\section{Introduction}
Machine translation is a task that aims to translate a sentence from one language to another language. Most of the proposed methods for machine translation models are based on encoder-decoder models \cite{sutskever2014sequence} \cite{bahdanau2014neural} \cite{luong2015effective} \cite{vaswani2017attention}. The encoder transforms the source language sentence into a vector which is used in the decoder to generate the target language sentence. Recently, with the development of pre-trained language models that make significant improvements in many NLP tasks. Especially, in the machine translation task, multilingual pre-trained models like mBART \cite{liu2020multilingual} M2M \cite{fan2021beyond} have been pre-training in many languages and giant datasets.

In Vietnamese, spite neural machine translation(NMT) achieves impressive results in English-Vietnamese \cite{doan2021phomt} \cite{ngo2022mtet} which were trained on the large and high-quality parallel corpus. However, in low-resource pairs language, NMT faces a lack of datasets such as in Vietnamese-Chinese, Vietnamese-France, etc. Building a large and high-quality parallel dataset is quite difficult and takes a lot of cost and time while monolingual corpora are available on the internet. 

Therefore, in this paper, we propose an efficient approach for machine translation on low-resource languages. Our system participates in the Vietnamese-Chinese Machine Translation challenge in VLSP 2022. The main works can be summarized as follows:
\begin{enumerate}
    \item We employ the strong multilingual sequence to sequence the pre-trained language model mBART for the machine translation task.
    \item We propose data selection and data synthesis techniques from the monolingual corpus that helps improve our machine translation system. 
\end{enumerate}
 The rest of this paper is organized as follows. We first review related work on machine translation methods and recent approaches to low-resource languages. We then present our system, including the model architecture and data processing techniques. Experimental results and analyses are described in the next section. Finally, we conclude the paper. 

\section{Related Work}
This section gives a brief review of related work on deep learning-based models for the machine translation task and recent solutions for low-resource languages.

Neural machine translation(NMT) is an approach to machine translation that was proposed first time by \cite{kalchbrenner-blunsom-2013-recurrent} and \cite{sutskever2014sequence} which only have two simple components including encoder and decoder based on Gated Recurrent Units (GRU) \cite{chung2014empirical} or Long Short-Term Memory (LSTM) \cite{hochreiter1997long}. This neural network compresses all the information of a source language sentence into a fixed-length vector. This may make it difficult for the neural network to translate long sentences. To deal with this problem \cite{bahdanau2014neural} and \cite{luong2015effective} proposed attention mechanisms that use alignment information between a source sentence and the corresponding target sentence to enhance the effectiveness of the systems. The transformers architecture is presented first time for machine translation by \cite{vaswani2017attention}. They proposed a self-attention mechanism that is a variant of the original attention \cite{bahdanau2014neural} and \cite{luong2015effective}. By using self-attention, a word is attended to the other words in a sentence. In recent years, with an explosion of deep learning, transformer architectures  for language models such as BERT \cite{devlin2018bert}, M2M \cite{fan2021beyond}, mBART \cite{liu2020multilingual} is capable of contextual representation of sentences. Based on the contextual representation ability, the machine translation task has gained significant improvements. 

Machine translation in low-resource languages is the most difficult problem. To deal with this problem, the most popular method is using the synthesis data from monolingual corpora to enhance the performance of the NMT system. Especially, in Vietnamese,  \cite{ngo-etal-2020-improving}, \cite{thi-vinh-etal-2020-uet}, \cite{cuong-thu-2020-vietnamese}, \cite{ngo2022efficient} proposed method using back-translation to generate parallel dataset such as Vietnamese-English, Vietnamese-Chinese, Vietnamese-France, Vietnamese-Japanese. Following these works, we also use back-translation as a data augmentation method but have some adjustments. 

\section{Approach}
\subsection{Our proposed method}
\begin{figure*}[t]

    \centering
    \includegraphics[width=1\textwidth]{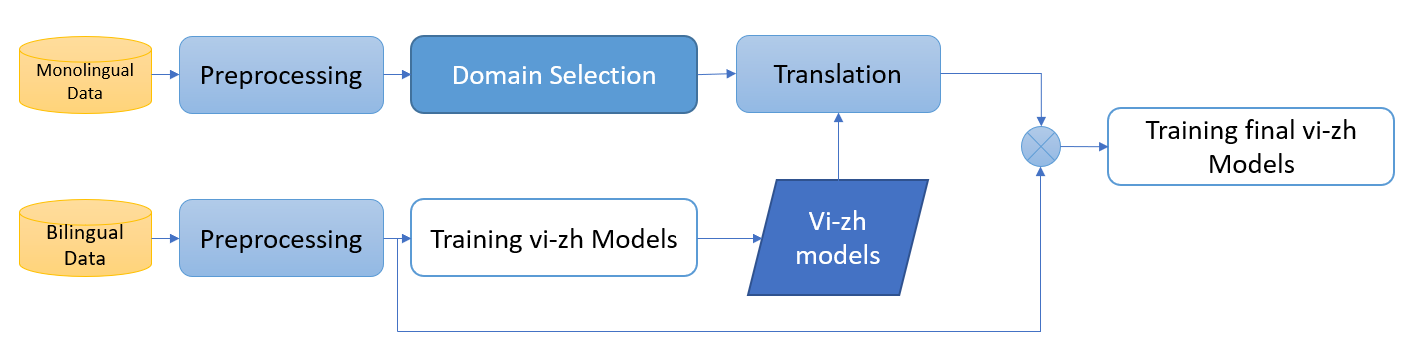}
    \caption{Flow of data processing and model training}
    \label{fig:translate_flow}
\end{figure*}

For the low-resource bilingual dataset like Vietnamese-Chinese, we proposed to use data synthesis as an augmentation technique to build more data for the training corpus. Firstly, an machine translation system is finetuned from the pre-trained mBart-50 using existing parallel data. This system is used to translate the target to the source language. The result is a new parallel corpus in which the source side is 200k sentences that are cleaned and selected from nearly 20M sentences for each language (monolingual dataset) by TF-IDF. Then, the synthesized parallel corpus is combined with the real text (bilingual dataset) to train the final system. Figure \ref{fig:translate_flow} illustrates our proposed training flow. In this paper, we adopted mBart-50 as the main translation model. Both monolingual and bilingual datasets must be cleaned and preprocessed before feeding to the mBart-50. To build the final translation model, three main steps have to be performed:
\begin{itemize}
    \item Step 1: Training a Vietnamese-English translation model with mBart-50 using the bilingual dataset.
    \item Step 2: Generating an extra bilingual dataset from the clean and selected sentences from the monolingual dataset.
    \item Step 3: Finetuning translation model with the newly generated extra bilingual dataset.
\end{itemize}

We used the same flow for the Chinese-Vietnamese or Vietnamese-Chinese translation models.


\subsection{Text Pre-processing}
\subsubsection{Data cleaning and Preprocessing}
The dataset given by 2022 VLSP organizers for the Vietnamese-Chinese Machine Translation task includes Parallel Corpora which has 300k sentence pairs and Monolingual Corpora which has 25M Vietnamese sentences and 19M Chinese sentences. Because of the noisy data, some steps have been taken to clean this dataset.
\begin{itemize}
    \item Removing non-Vietnamese and non-Chinese sentences using a language detection model pycld3 \footnote{https://pypi.org/project/pycld3}.
    \item Cleaning HTML characters and some special characters.
    \item Removing sentences that are too long (more than 100 words) or too short (less than 2 words).
\end{itemize}

\subsection{Mbart model for Neural Machine Translation}
For low-resource pair languages like Vietnamese-Chinese or Vietnamese-Korean, public datasets for training NMT are limited both in quantity and quality. As a result, training an NMT model based on a Transformer from scratch led to low accuracy \cite{koehn-knowles-2017-six}. Besides, mBart demonstrates that multilingual
denoising pre-training produces significant performance improvement across a wide variety of
machine translation (MT) tasks \cite{DBLP:journals/corr/abs-2001-08210}. Therefore, we employ the mBart for the machine translation task.
mBART-50 is a multilingual Sequence-to-Sequence model. It was introduced to show that multilingual translation models can be created through multilingual fine-tuning. Instead of fine-tuning in one direction, a pre-trained model is fine-tuned in many directions simultaneously. mBART-50 is created using the original mBART model and extended to add extra 25 languages to support multilingual machine translation models of 50 languages as mentioned in \cite{DBLP:journals/corr/abs-2008-00401}. The model is already trained in Vietnamese and Chinese. Subsequently, we only need finetuning with the parallel data from VLSP to have the baseline models.
\subsubsection{Data selection and Parallel data synthesis}
To utilize monolingual corpora, a data selection technique is applied to both 25M Vietnamese sentences and 19M Chinese sentences. TF-IDF scores for each sentence are calculated and ranked to filter out 200k sentences in the training data's domain for each language. Then, baseline NMT models translate the source language to the target language to have new synthesized parallel data, which are combined with the original parallel data for training the final model.
\section{Experiments}
\subsection{Experimental Setup}
Our models were implemented in Pytorch using HuggingFace\footnote{https://huggingface.co}. We used  mBART50 model\footnote{https://huggingface.co/facebook/mbart-large-50} as pre-trained language models. In all experiments, we set the max sequence length to $100$ and used the beam search with a beam size was 4. We trained our models using the AdamW optimizer \cite{Loshchilov:2019} with a batch size of $16$. We set the epsilon and weight decay to the default value in PyTorch, i.e., 1e-8. The learning rate was tuned in \{1e-5, 2e-5, 3e-5, 4e-5, 5e-5\}. For each model, we trained for $5$ epochs and calculated the BLEU score after each epoch on the validation set. The version with the highest BLEU score was selected to apply to the public test set. Our model's hyperparameters are summarized in Table \ref{tab:params}. 
\begin{table}[t]
\centering
\caption{Hyperparameters of our model}
\label{tab:params}
\begin{tabular}{|l|c|}
\hline
\textbf{Hyperprameter}	& \textbf{Value}\\
\hline
\hline
Max sequence length	& 100\\
\hline
Batch size	& 16\\
\hline
Epochs	& 4\\

\hline
Learning rate &	4e-5\\
\hline
Weight decay &	1e-8\\
\hline
Beam size  &	4\\
\hline
\end{tabular}
\end{table}

\subsection{Experiemtal Results}
\begin{table*}[t]
\centering
\caption{The main results in Vietnamese to Chinese translation}
\begin{tabular}{|l|c|c|}
\hline\textbf{Systems}& \textbf{dev} & \textbf{test} \\
\hline Google Translate & \textbf{60.1} &	\textbf{41.6} \\
\hline UET Engine\footnote{https://dichmay.itrithuc.vn/} & 41.8 &	39.2 \\
\hline Our system training on original VLSP 2022 dataset & 33.10 & 38.22  \\
\hline + Domain data selection & 34.03 & 38.13 \\
\hline + train from scratch ( original dataset + ranked bilingual synthesis dataset) & 34.58 & 38.78  \\
\hline + fine-tuning (original dataset + ranked bilingual synthesis dataset) & \underline{36.76} &	\underline{38.97} \\
\hline
\end{tabular}
\label{tab:vi-cn}
\end{table*}

\begin{table*}[t]
\centering
\caption{The main results in Chinese to Vietnamese translation}
\begin{tabular}{|l|c|c|}
\hline\textbf{Systems}& \textbf{dev} & \textbf{test} \\
\hline Google Translate & 38.5 &	38.1 \\
\hline UET Engine & \textbf{43.3} &	\textbf{40.6} \\
\hline Our system training on original VLSP 2022 dataset & 33.23 & 35.58  \\
\hline + Domain data selection & 33.05 & 36.01 \\
\hline + train from scratch ( original dataset + ranked bilingual synthesis dataset) & \underline{35.50} & \underline{38.90}  \\
\hline + fine-tuning (original dataset + ranked bilingual synthesis dataset) & 34.92 &	38.77 \\
\hline
\end{tabular}
\label{tab:cn-vi}
\end{table*}

From table \ref{tab:cn-vi} and \ref{tab:vi-cn}, the result shows a significant improvement with a 3.19 BLEU score on the test set for Chinese to Vietnamese direction while a small gain on BLEU score on the test set for Vietnamese to Chinese when monolingual corpora and domain selection techniques are applied. Especially, our model has a score higher than 0.67 compared to Google Translate in Chinese to Vietnamese translation. Overall, Google Translate is the best for Vietnamese-to-Chinese translation while the UET engine is the best for Chinese-to-Vietnamese translation.

\section{Conclusion}
Our approach has achieved significant improvements by leveraging the multilingual sequence-to-sequence model mBART and data synthesis. Despite the size of the bilingual corpus is limited, the results of both translation directions are comparable to other engines even with a small parallel dataset. In the future, we will investigate other techniques such as contrastive learning for machine translation to improve our system.
\bibliography{vlsp2020}

\begin{thebibliography}{20}
\expandafter\ifx\csname natexlab\endcsname\relax\def\natexlab#1{#1}\fi

\bibitem[{Bahdanau et~al.(2014)Bahdanau, Cho, and Bengio}]{bahdanau2014neural}
Dzmitry Bahdanau, Kyunghyun Cho, and Yoshua Bengio. 2014.
\newblock Neural machine translation by jointly learning to align and
  translate.
\newblock \emph{arXiv preprint arXiv:1409.0473}.

\bibitem[{Chung et~al.(2014)Chung, Gulcehre, Cho, and
  Bengio}]{chung2014empirical}
Junyoung Chung, Caglar Gulcehre, KyungHyun Cho, and Yoshua Bengio. 2014.
\newblock Empirical evaluation of gated recurrent neural networks on sequence
  modeling.
\newblock \emph{arXiv preprint arXiv:1412.3555}.

\bibitem[{Cuong and Thu(2020)}]{cuong-thu-2020-vietnamese}
Le~Duc Cuong and Trang Nguyen~Thi Thu. 2020.
\newblock \href {https://aclanthology.org/2020.vlsp-1.13}
  {{V}ietnamese-{E}nglish translation with transformer and back translation in
  {VLSP} 2020 machine translation shared task}.
\newblock In \emph{Proceedings of the 7th International Workshop on Vietnamese
  Language and Speech Processing}, pages 64--70, Hanoi, Vietnam. Association
  for Computational Lingustics.

\bibitem[{Devlin et~al.(2018)Devlin, Chang, Lee, and
  Toutanova}]{devlin2018bert}
Jacob Devlin, Ming-Wei Chang, Kenton Lee, and Kristina Toutanova. 2018.
\newblock Bert: Pre-training of deep bidirectional transformers for language
  understanding.
\newblock \emph{arXiv preprint arXiv:1810.04805}.

\bibitem[{Doan et~al.(2021)Doan, Nguyen, Tran, Hoang, and
  Nguyen}]{doan2021phomt}
Long Doan, Linh~The Nguyen, Nguyen~Luong Tran, Thai Hoang, and Dat~Quoc Nguyen.
  2021.
\newblock Phomt: A high-quality and large-scale benchmark dataset for
  vietnamese-english machine translation.
\newblock \emph{arXiv preprint arXiv:2110.12199}.

\bibitem[{Fan et~al.(2021)Fan, Bhosale, Schwenk, Ma, El-Kishky, Goyal, Baines,
  Celebi, Wenzek, Chaudhary et~al.}]{fan2021beyond}
Angela Fan, Shruti Bhosale, Holger Schwenk, Zhiyi Ma, Ahmed El-Kishky,
  Siddharth Goyal, Mandeep Baines, Onur Celebi, Guillaume Wenzek, Vishrav
  Chaudhary, et~al. 2021.
\newblock Beyond english-centric multilingual machine translation.
\newblock \emph{J. Mach. Learn. Res.}, 22(107):1--48.

\bibitem[{Hochreiter and Schmidhuber(1997)}]{hochreiter1997long}
Sepp Hochreiter and J{\"u}rgen Schmidhuber. 1997.
\newblock Long short-term memory.
\newblock \emph{Neural computation}, 9(8):1735--1780.

\bibitem[{Kalchbrenner and Blunsom(2013)}]{kalchbrenner-blunsom-2013-recurrent}
Nal Kalchbrenner and Phil Blunsom. 2013.
\newblock \href {https://aclanthology.org/D13-1176} {Recurrent continuous
  translation models}.
\newblock In \emph{Proceedings of the 2013 Conference on Empirical Methods in
  Natural Language Processing}, pages 1700--1709, Seattle, Washington, USA.
  Association for Computational Linguistics.

\bibitem[{Koehn and Knowles(2017)}]{koehn-knowles-2017-six}
Philipp Koehn and Rebecca Knowles. 2017.
\newblock \href {https://doi.org/10.18653/v1/W17-3204} {Six challenges for
  neural machine translation}.
\newblock In \emph{Proceedings of the First Workshop on Neural Machine
  Translation}, pages 28--39, Vancouver. Association for Computational
  Linguistics.

\bibitem[{Liu et~al.(2020{\natexlab{a}})Liu, Gu, Goyal, Li, Edunov,
  Ghazvininejad, Lewis, and Zettlemoyer}]{liu2020multilingual}
Yinhan Liu, Jiatao Gu, Naman Goyal, Xian Li, Sergey Edunov, Marjan
  Ghazvininejad, Mike Lewis, and Luke Zettlemoyer. 2020{\natexlab{a}}.
\newblock Multilingual denoising pre-training for neural machine translation.
\newblock \emph{Transactions of the Association for Computational Linguistics},
  8:726--742.

\bibitem[{Liu et~al.(2020{\natexlab{b}})Liu, Gu, Goyal, Li, Edunov,
  Ghazvininejad, Lewis, and Zettlemoyer}]{DBLP:journals/corr/abs-2001-08210}
Yinhan Liu, Jiatao Gu, Naman Goyal, Xian Li, Sergey Edunov, Marjan
  Ghazvininejad, Mike Lewis, and Luke Zettlemoyer. 2020{\natexlab{b}}.
\newblock \href {http://arxiv.org/abs/2001.08210} {Multilingual denoising
  pre-training for neural machine translation}.
\newblock \emph{CoRR}, abs/2001.08210.

\bibitem[{Loshchilov and Hutter(2019)}]{Loshchilov:2019}
Ilya Loshchilov and Frank Hutter. 2019.
\newblock Decoupled weight decay regularization.
\newblock In \emph{Proceedings of the International Conference on Learning
  Representations (ICLR)}.

\bibitem[{Luong et~al.(2015)Luong, Pham, and Manning}]{luong2015effective}
Minh-Thang Luong, Hieu Pham, and Christopher~D Manning. 2015.
\newblock Effective approaches to attention-based neural machine translation.
\newblock \emph{arXiv preprint arXiv:1508.04025}.

\bibitem[{Ngo et~al.(2022{\natexlab{a}})Ngo, Trinh, Phan, Tran, Dang, Nguyen,
  Nguyen, and Luong}]{ngo2022mtet}
Chinh Ngo, Trieu~H Trinh, Long Phan, Hieu Tran, Tai Dang, Hieu Nguyen, Minh
  Nguyen, and Minh-Thang Luong. 2022{\natexlab{a}}.
\newblock Mtet: Multi-domain translation for english and vietnamese.
\newblock \emph{arXiv preprint arXiv:2210.05610}.

\bibitem[{Ngo et~al.(2020)Ngo, Nguyen, Ha, Dinh, and
  Nguyen}]{ngo-etal-2020-improving}
Thi-Vinh Ngo, Phuong-Thai Nguyen, Thanh-Le Ha, Khac-Quy Dinh, and Le-Minh
  Nguyen. 2020.
\newblock \href {https://www.aclweb.org/anthology/2020.loresmt-1.8} {Improving
  multilingual neural machine translation for low-resource languages: {F}rench,
  {E}nglish - {V}ietnamese}.
\newblock In \emph{Proceedings of the 3rd Workshop on Technologies for MT of
  Low Resource Languages}, pages 55--61, Suzhou, China. Association for
  Computational Linguistics.

\bibitem[{Ngo et~al.(2022{\natexlab{b}})Ngo, Nguyen, Nguyen, Ha, and
  Nguyen}]{ngo2022efficient}
Thi-Vinh Ngo, Phuong-Thai Nguyen, Van~Vinh Nguyen, Thanh-Le Ha, and Le-Minh
  Nguyen. 2022{\natexlab{b}}.
\newblock An efficient method for generating synthetic data for low-resource
  machine translation: An empirical study of chinese, japanese to vietnamese
  neural machine translation.
\newblock \emph{Applied Artificial Intelligence}, 36(1):2101755.

\bibitem[{Sutskever et~al.(2014)Sutskever, Vinyals, and
  Le}]{sutskever2014sequence}
Ilya Sutskever, Oriol Vinyals, and Quoc~V Le. 2014.
\newblock Sequence to sequence learning with neural networks.
\newblock \emph{Advances in neural information processing systems}, 27.

\bibitem[{Tang et~al.(2020)Tang, Tran, Li, Chen, Goyal, Chaudhary, Gu, and
  Fan}]{DBLP:journals/corr/abs-2008-00401}
Yuqing Tang, Chau Tran, Xian Li, Peng{-}Jen Chen, Naman Goyal, Vishrav
  Chaudhary, Jiatao Gu, and Angela Fan. 2020.
\newblock \href {http://arxiv.org/abs/2008.00401} {Multilingual translation
  with extensible multilingual pretraining and finetuning}.
\newblock \emph{CoRR}, abs/2008.00401.

\bibitem[{Thi-Vinh et~al.(2020)Thi-Vinh, Minh-Thuan, Cong, Hoang-Quan,
  Phuong-Thai, and Van-Vinh}]{thi-vinh-etal-2020-uet}
Ngo Thi-Vinh, Nguyen Minh-Thuan, Nguyen Hoang~Minh Cong, Nguyen Hoang-Quan,
  Nguyen Phuong-Thai, and Nguyen Van-Vinh. 2020.
\newblock \href {https://aclanthology.org/2020.vlsp-1.14} {The {UET}-{ICTU}
  submissions to the {VLSP} 2020 news translation task}.
\newblock In \emph{Proceedings of the 7th International Workshop on Vietnamese
  Language and Speech Processing}, pages 71--76, Hanoi, Vietnam. Association
  for Computational Lingustics.

\bibitem[{Vaswani et~al.(2017)Vaswani, Shazeer, Parmar, Uszkoreit, Jones,
  Gomez, Kaiser, and Polosukhin}]{vaswani2017attention}
Ashish Vaswani, Noam Shazeer, Niki Parmar, Jakob Uszkoreit, Llion Jones,
  Aidan~N Gomez, {\L}ukasz Kaiser, and Illia Polosukhin. 2017.
\newblock Attention is all you need.
\newblock \emph{Advances in neural information processing systems}, 30.

\end{thebibliography}
\bibliographystyle{acl_natbib}

\end{document}